\documentclass[letterpaper, 10 pt, conference]{ieeeconf}  

\IEEEoverridecommandlockouts                              

\usepackage{listings} 
\usepackage{xcolor}
\usepackage{graphicx}
\usepackage{amsmath}
\usepackage{subcaption}
\usepackage{booktabs}
\usepackage{multirow} 
\usepackage{multicol} 
\usepackage{xspace} 
\usepackage{verbatim} 
\usepackage{siunitx} 
\usepackage{hyperref} 

\usepackage[style=ieee,sorting=nyt,maxcitenames=4,mincitenames=1,maxbibnames=4]{biblatex}
\usepackage[font=small,labelfont=bf]{caption}  

\newcommand{\method}{IMPACT\xspace}
\newcommand{\baseline}{VLM Cost\xspace}

\newcommand{\remark}[3]{{\color{#2}[#1: #3]}}
\newcommand{\daniel}[1]{\remark{Daniel}{cyan}{#1}} 
\newcommand{\yiyang}[1]{\remark{Yiyang}{green}{#1}}

\definecolor{lightblue}{RGB}{0, 128, 255}

\newcommand{\ie}{i.e.,\xspace}
\newcommand{\eg}{e.g.,\xspace}

\usepackage{listings}
\usepackage{fancyhdr}
\title{\LARGE \bf
IMPACT\includegraphics[height=0.80em]{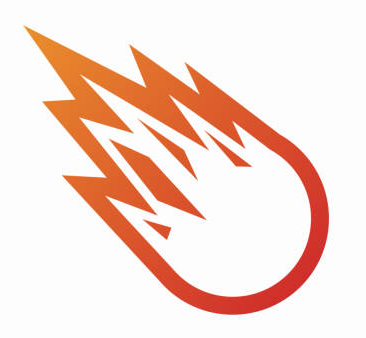}: Intelligent Motion Planning with\\Acceptable Contact Trajectories via Vision-Language Models
}

\author{Yiyang Ling$^{*}$, Karan Owalekar$^{*}$, Oluwatobiloba Adesanya, Erdem Bıyık, Daniel Seita
\thanks{*Equal contribution.}
\thanks{All authors are with Thomas Lord Department of Computer Science, Viterbi School of Engineering, University of Southern California, USA.}
\thanks{Correspondence: {\tt\small \{lingyiya,kowaleka,seita\}@usc.edu}.}%
}

\begin{document}

\maketitle

\begin{abstract}
Motion planning involves determining a sequence of robot configurations to reach a desired pose, subject to movement and safety constraints. 
Traditional motion planning finds collision-free paths, but this is overly restrictive in clutter, where it may not be possible for a robot to accomplish a task without contact. 
In addition, contacts range from relatively benign (\eg brushing a soft pillow) to more dangerous (\eg toppling a glass vase), making it difficult to characterize which may be acceptable. In this paper, we propose \method, a novel motion planning framework that uses Vision-Language Models (VLMs) to infer environment semantics, identifying which parts of the environment can best tolerate contact based on object properties and locations. 
Our approach generates an anisotropic cost map that encodes directional push safety. We pair this map with a contact-aware A* planner to find stable contact-rich paths.
We perform experiments using 20 simulation and 10 real-world scenes and assess using task success rate, object displacements, and feedback from human evaluators. 
Our results over 3200 simulation and 200 real-world trials suggest that \method enables efficient contact-rich motion planning in cluttered settings while outperforming alternative methods and ablations. Our project website is available at \href{https://impact-planning.github.io/}{https://impact-planning.github.io/}.
\end{abstract}

\section{Introduction}
\label{sec:intro}

Classical motion planning for robot manipulation~\cite{modern_robotics_2017,manipulation} frames the problem as finding a path for the robot's end-effector to reach a target while avoiding collisions with obstacles. 
This formulation is generally desirable, but can be highly restrictive, especially in densely cluttered environments~\cite{dogar2011frameworkpushgrasp}. In such cases, some incidental contact may be necessary to achieve a task, or to accomplish it more efficiently than by strictly avoiding all collisions. 

\begin{figure}[t]
    \centering
    \includegraphics[width=0.48\textwidth]{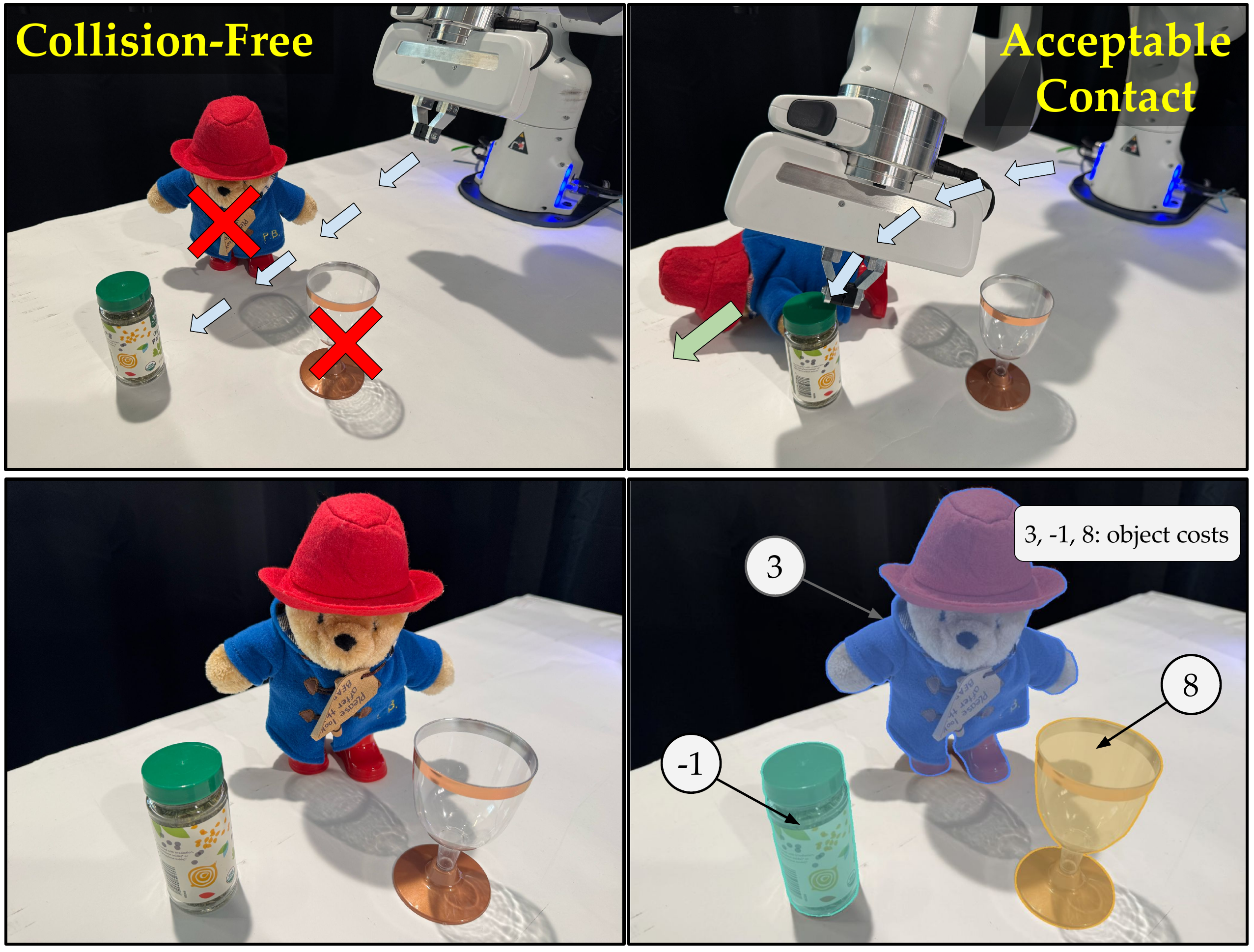}
    \caption{
        An example of a reaching task and object costs. The \underline{first} row shows the difference between collision-free paths and paths with acceptable contact. Left: Collision-free paths prevent a “straight” path to the spice jar because of the toy bear and wine glass obstacles (each marked with a red ``X''). Right: With semantically acceptable contact, the robot can successfully reach the spice jar by pushing the toy bear and avoiding the fragile wine glass. 
        The \underline{second} row shows the cost of each object generated by GPT-4o. Left: the original scene. Right: GPT-4o assigns different costs to objects, with the target assigned $-1$ (toy bear: $3$, spice jar: $-1$, wine glass: $8$).}
    \label{fig:example}
    \vspace{-12pt}
\end{figure}

Consider the example in Fig.~\ref{fig:example}, which shows a robot manipulator making contact with a toy bear to efficiently reach the target spice jar. Due to the clutter, a collision-free path either does not exist or would require a longer parabolic motion to go above the obstacles (which, again, may not be feasible in cluttered cabinets and boxes). 
We thus desire robots that can achieve a task while moving through a cluttered environment, making appropriate contact as needed. 
In this work, the term ``contacts'' does not refer to when a robot uses its grippers to touch a target object (\eg for grasping). Instead, we consider other types of contact: when any part of the robot touches any \emph{non-target} (or ``distractor'') object in the environment. We study this in motion planning for tasks that involve reaching to a target in dense clutter. 

To address this challenge, we propose \textbf{I}ntelligent  \textbf{M}otion  \textbf{P}lanning with  \textbf{A}cceptable  \textbf{C}ontact  \textbf{T}rajectories (\method), a novel framework that leverages modern Vision-Language Models (VLMs) such as GPT-4o~\cite{openai2024gpt4ocard} to infer initial semantic costs for objects in the scene. However, knowing that an object is generally safe to touch is different from knowing \textit{how} and \textit{where} to push it without causing a cascade of collisions.
Our method repeatedly samples push outcomes to estimate the probability of causing a collision with obstacles or the target. This risk is condensed into safety scores that are used to construct an anisotropic cost map, where interaction costs vary with direction since pushing an object from one side versus another can lead to different outcomes. We integrate this directional cost map with a contact-aware A* planner~\cite{hart1968formal} that computes paths to reach targets.
We perform experiments on contact-rich, densely cluttered reaching tasks in simulation and real-world settings. We evaluate \method using multiple quantitative metrics, including human preference rankings from a user study. Results indicate that users prefer \method over alternatives, suggesting that it is a promising approach for facilitating semantically-acceptable contact-rich manipulation.

Our contributions are as follows:
\begin{itemize}
\item \method, a framework that formalizes ``acceptable contact'' by transforming initial VLM-inferred semantic costs into a dense, anisotropic cost map to represent the directional safety of physical interactions.
\item A contact-aware A* planner capable of interpreting this anisotropic map to execute paths with intelligent and minimal-impact contact.
\item Extensive simulation and real-world experiments, including a human subjects study, that validate our approach using both objective and subjective metrics to evaluate the success of contact-rich trajectories.
\end{itemize}


\section{Related Work}
\label{sec:related}

\subsection{Robot Motion Planning}


Classical motion planning algorithms can be broadly characterized~\cite{manipulation} as optimization-based such as TrajOpt~\cite{schulman2013finding}, or sampling-based such as PRM~\cite{PRM}, RRT~\cite{lavalle1998rapidly}, and RRT*~\cite{karaman2011sampling}. 
While there are numerous variants of these methods, a common underlying theme is the constraint of avoiding any collisions. 
In densely cluttered scenarios where contact is inevitable, these motion planners might not find a solution, even though one might exist if the robot engages in light contact. 
Prior work like Navigation Among Movable Obstacles~\cite{stilman2005navigation} and Minimum Constraint Removal~\cite{hauser2013minimum, thomas2023computational} reason about relocating or removing obstacles to obtain feasible paths.
Our method seamlessly integrates with prior motion planning algorithms for flexible contact-rich manipulation, without requiring explicit object relocation or environment modification.

Some motion planning works modify cost functions so that certain obstacles, such as leaves in foliage, are permeable~\cite{killpack2016model,nemlekar2021robotic} and thus allow for some contact. These methods have been applied to domain-specific foliage settings and not tested in common cluttered scenarios with rigid objects. 
In closely-related work, Xie~et~al.~\cite{xielanguage} propose to use semantic language commands to enable semantically-acceptable contact. Unlike~\cite{xielanguage}, we do not require explicit language instructions about which contacts are acceptable or not, since we utilize VLMs to automate this process. 

\subsection{Contact-Rich Robot Manipulation}

Dealing with contacts is challenging in manipulation~\cite{manipulation}. 
This may refer to frequent contact between an object (that the robot grips/holds) and an environment, such as tight placement tasks like connector and peg insertion as explored in robotic reinforcement learning works~\cite{levine_finn_2016,DeepRL_insertion_2020,multimodal_2019}. 
Another method for contact-rich problems is extrinsic dexterity~\cite{Chavan-Dafle_2014,Wenxuan_extrinsic_2022}, which takes advantage of contacts between an object and rigid parts of the environment (such as walls) to reorient the object to improve subsequent manipulation. 
Other works consider contact-rich interactions in different applications such as assistive robots that incorporate human feedback and touch~\cite{robot_bite_transfer_2024}. 
In contrast, we consider the relatively less-explored ``contact-rich'' setting, where ``contact'' refers to robot parts touching the environment's obstacles.

\subsection{Vision-Language Models (VLMs) for Robotics}

VLMs such as GPT-4o~\cite{openai2024gpt4ocard} and Gemini~\cite{Gemini} are increasingly applied in robotics for high-level planning~\cite{ahn2022can,driess2023palmeembodiedmultimodallanguage,liang2023code,singh2023progprompt}, reward or task design ~\cite{maeureka,sontakke2023roboclip,wanggensim,RL-VLM-F},  and low-level affordance reasoning~\cite{kuang2024ramretrievalbasedaffordancetransfer,liu2024moka}.
In contrast to these works, we use VLMs specifically to infer object contact tolerances to guide contact-rich motion planning. 

In closely-related work, VoxPoser~\cite{huang2023voxposer} exploits VLMs for open-world reasoning and visual grounding to compose a 3D value map to guide robotic interactions.
As in~\cite{huang2023voxposer}, we use VLMs to assign semantic costs to objects. However, unlike VoxPoser, which requires the user to explicitly specify which objects to avoid with language, \eg \emph{``get the item, but watch out for that vase!''} we do not require explicit language commands. 
Instead, we leverage the improving spatial and semantic knowledge of recent VLMs~\cite{PhysBench} to determine object contact tolerances, and show manipulation in more densely cluttered environments. 
Other recent work relies on VLMs for semantically-safe manipulation~\cite{brunke2024semanticallysaferobotmanipulation} but assumes that any collisions are undesirable. 


\section{Problem Statement and Assumptions}
\label{sec:problem}

We assume a single robot arm with a standard gripper operates in a densely cluttered environment that contains $n$ objects, denoted as ${\mathcal{O} = \{o_1, o_2, \ldots, o_n\}}$. The robot must reach a given target object ${o_{\rm targ} \in \mathcal{O}}$ while minimizing unwanted contact with other objects (\ie obstacles) in ${\mathcal{O} \setminus o_{\rm targ}}$.  
In environments with significant clutter, $o_{\rm targ}$ may be behind or close to multiple objects and thus reaching it may be infeasible with a collision-free trajectory. 
To reduce occlusions, at least two cameras provide respective RGBD images at the start of the task. 
Given these image observations, the objective is to compute a trajectory $\tau$ for the robot, defined as a sequence of end-effector poses, such that its end-effector ultimately touches $o_{\rm targ}$. 
Thus, among the dynamically feasible trajectories, our objective is to select one that reaches the target while engaging in ``semantically-acceptable'' contact with obstacles when needed.


\section{Method: \method}
\label{sec:method}

\begin{figure*}[t]
    \centering
    \includegraphics[width=\textwidth]{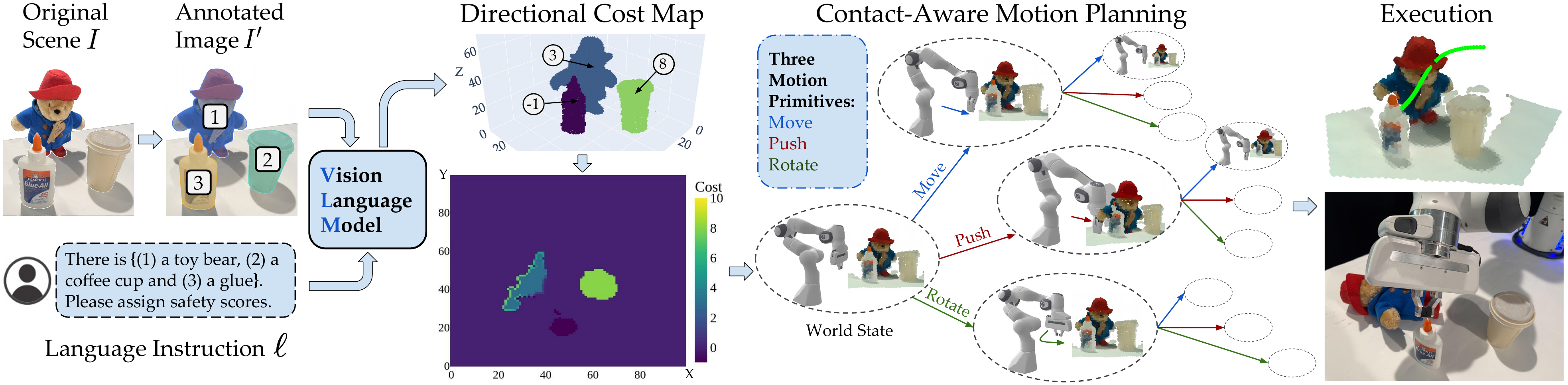} 
    \caption{
        Overview of \method. There is a toy bear, a coffee cup and a glue bottle (target) on the table. 
        The VLM receives an annotated image $I'$ and a language template prompt $\ell$ with object information from SAM2~\cite{ravi2024sam}, and outputs \emph{costs} for the three objects.
        We use a cost of $-1$ for the target object. We construct a 3D voxel grid $V$ using these costs and then flatten it to produce an anisotropic, contact-aware cost map $M'$. The contact-aware A* planner searches over three motion primitives in this map: \emph{Move}, \emph{Push} and \emph{Rotate} to generate a trajectory. The planner's state space includes the robot's end-effector pose and the displaced positions of low-cost objects. 
        These guide the robot to avoid the coffee cup but make contact with the toy bear at the appropriate direction to reach the glue bottle. 
        }
    \label{fig:method}
    \vspace{-12pt}
\end{figure*}

Our framework consists of two main steps (see Fig.~\ref{fig:method}). First, it uses a VLM to obtain object costs in a cluttered scene (Sec.~\ref{ssec:object_costs}), and then uses those costs for contact-rich motion planning (Sec.~\ref{ssec:motion_planning_contacts}).

\subsection{Obtaining Object Costs using a VLM}
\label{ssec:object_costs}

A key technical challenge is defining the notion of an ``acceptable'' contact. This depends heavily on semantics, or the general-purpose commonsense knowledge that humans have about the behavior of diverse objects. Different objects with varying materials, geometries, sizes or purposes have different tolerances to potential contact. Furthermore, tolerance to contact should also depend on an object's proximity to nearby objects. Therefore, we leverage the commonsense knowledge in VLMs to estimate the tolerance rate for contact of each object. 
We encode this information by assigning each object to an integer in ${\{0, 1, \ldots, 10\}}$ as the cost, where a higher cost indicates lower tolerance to contact with any part of the robot arm as it executes a trajectory.
For example, the cost of a fragile object (\eg wine glass) should be significantly higher compared to the cost of an object that can better absorb contact (\eg foam rubber). 

We use GPT-4o (hereafter, GPT) as the VLM to generate the cost of all objects in the scene due to its strong spatial reasoning capabilities~\cite{PhysBench}, but our approach is compatible with other VLMs. 
The VLM is used in a zero-shot inference setting without any fine-tuning.
The input to the VLM includes both an annotated image and a text prompt. 
We mount an RGBD camera to capture a scene image $I$, and use SAM2~\cite{ravi2024sam} to segment and annotate the objects.
The segmented and annotated image $I'$ is ultimately provided as part of the input to GPT. 
We also design a text prompt template $\ell$ which includes the list of objects in the scene (but is otherwise task-agnostic), and some general principles related to the concept of contact tolerance. Objects are labeled with the same numbers as in the input image, marked by SAM2.  
The full text prompt can be found on our website. The output of GPT is a dictionary with the cost of all queried objects. 
Fig.~\ref{fig:example} (second row) shows an example of object costs generated from GPT. It assigns a high cost of $8$ to the wine glass and a lower cost of $3$ to the toy bear, which indicates different tolerances to contact across objects.

\subsection{Directional Contact-Aware Motion Planning}
\label{ssec:motion_planning_contacts}

We use motion planning algorithms to synthesize robot trajectories. We construct a 3D voxel grid $V$ where each voxel $V[x,y,z]$ denotes the cost at position $(x,y,z)$. During planning, we assign all voxels corresponding to the target object a cost of $-1$ to encourage the planner to find a path to it. To evaluate directional contact information, we project $V$ into a 2D grid $M$ using a top-down view, where ${M[x,y] = \max_{z} V[x,y,z]}$. From this 2D perspective, all obstacles are divided into low-cost (5 or lower) and high-cost (6 or higher) sets based on their VLM-assigned costs. To simplify planning, we define three motion primitives in 2D space:  (i) \emph{Move}, which translates the robot to one of eight neighboring positions; (ii) \emph{Rotate}, which discretizes the end-effector orientation changes within $\pm 45^\circ$; and (iii) \emph{Push}, which moves the end-effector within a bounded radius while making contact with an object. We assume the first two change the end-effector's pose without affecting the environment. In contrast, the Push primitive denotes end-effector translations that involve some robot-object contact. 

\subsubsection{Anisotropic Cost Map Generation}
\label{sssec:cost_map_gen}
The map $M$ encodes object costs independent of motion direction. In clutter, however, contact safety depends strongly on the direction where the robot approaches an object. To address this, we construct a new \textbf{anisotropic cost map} $M'$ that augments each point with directional safety information. Defining $M'$ over both 2D positions and 2D directions would be high-dimensional and computationally expensive, so we keep $M'$ as a 2D function. 
For each point $(x,y)$ on the boundary of a low-cost object, we consider the \emph{reverse surface normal} vector, which points ``inwards'' towards the object. This gives the most natural and direct push for the object, but to consider different possible pushes, we sample $m$ push outcomes with small variations in distance and angle, weighted by a 2D Gaussian centered on that normal (so outcomes closer to that vector receive higher likelihood).
Each hypothesized outcome is evaluated on $M$ by checking whether the displaced object would overlap with other objects, and classified into four categories: safe, low-cost contact, high-cost contact or contact with the target. Push outcomes in the same category are assigned an identical safety score $u\in(0,1]$, with larger values indicating safer outcomes. The aggregated safety score at $(x,y)$ is:
\begin{equation}
\label{eq:safety_score}
f_s(x,y) = \frac{\sum_{i=1}^{m} l_i u_i}{\sum_{i=1}^{m} l_i},
\end{equation}
where $u_i$ encodes the safety level, and $l_i$ is the likelihood of the $i$-th push outcome. 
This score is then used to update the cost map at $(x,y)$, creating the final anisotropic cost map $M'$ where each point has an associated directional push score:
\begin{equation}
    M'[x,y] = \alpha M[x,y]+(1- \alpha) [10-10f_s(x,y)].
\end{equation}
Here, $\alpha\in (0,1)$ is a weight parameter to balance the object cost and the directional safety score $f_s \in [0, 1]$, and the constant 10 ensures the updated cost in $M'$ remains consistent with the original object cost range in $M$.

\subsubsection{Contact-Aware A* Planning}
\label{sssec:a_star_planning}

To generate a plan, we search for a sequence of primitives that guides the end-effector to the target while accounting for directional costs in $M'$. We formulate this as a graph search problem and use A*~\cite{hart1968formal}, which is well-suited for this setting and provides deterministic control compared to sampling-based algorithms. 
The planner's state is defined as $S = (p, r, \mathcal{D})$, where $(p,r)$ is the robot end-effector position and orientation in 2D space and $\mathcal{D}$ is a list that tracks the cumulative 2D displacement of low-cost objects. At each step, the planner expands nodes using the three primitives.
For Push, we estimate the object displacement using $d=D\cdot\big(\cos(\varphi)\big)^{\gamma}$, where  $D$ is the maximum displacement when pushing in the optimal direction, $\varphi$ is the angular deviation between the push direction and the object’s reverse surface normal at the contact point, and the decay parameter $\gamma$ is set to $1.3$.
\method extends standard A* search by incorporating world state changes caused by pushes.
Each valid Push creates a new branch in the search tree where an object has been permanently displaced to a new position. 
The resulting state, $S' = (p', r', \mathcal{D'})$, contains the updated robot pose and an updated world configuration.

\begin{figure}[t]
    \centering
    \includegraphics[width=\linewidth]{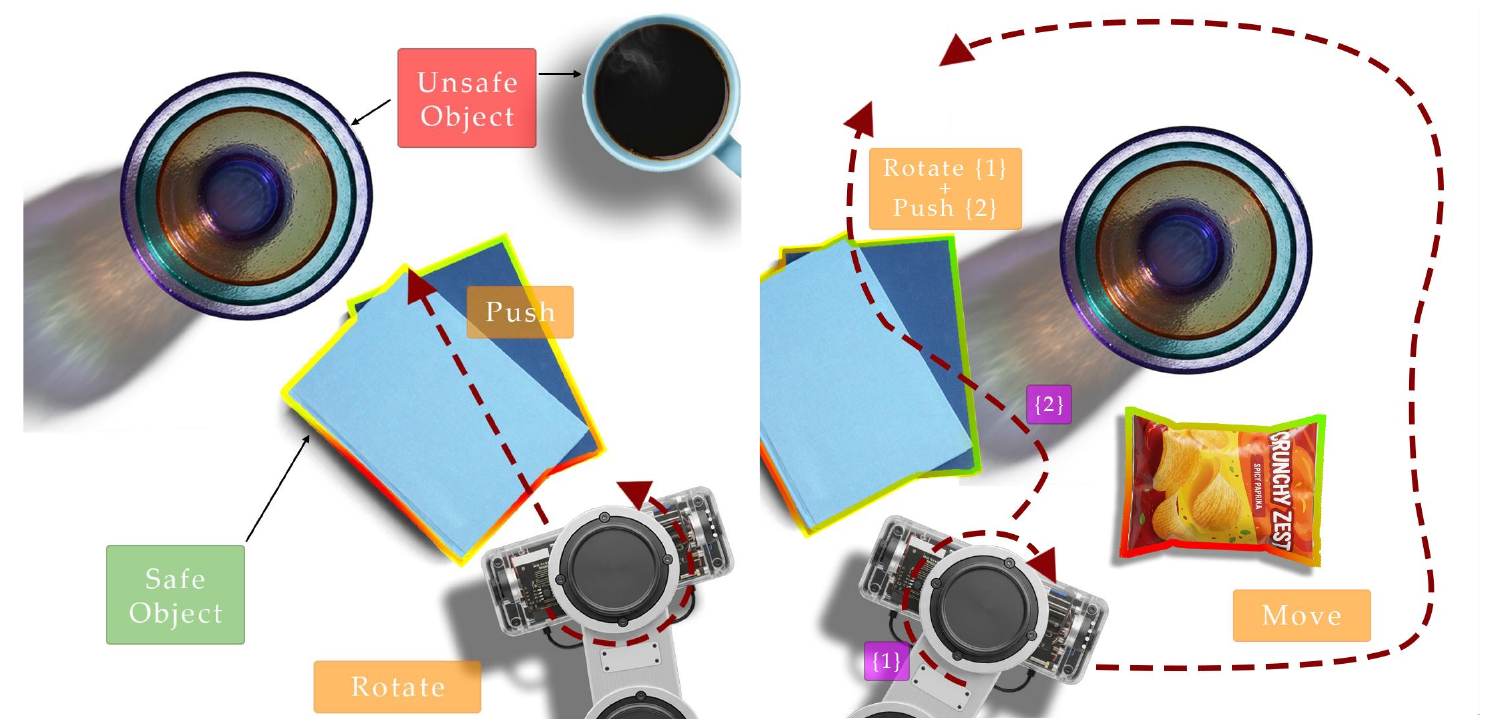}
    \caption{A* planner decision-making in two key scenarios, where the books and pack of chips are low-cost objects, while the mug and the stack of bowls are high-cost objects. The colored border around low cost objects visualizes anisotropic costs (red is unsafe, green is safe). 
    On the left, the planner avoids a direct Push towards the stack of bowls (cost $\infty$) and instead chooses a low-cost Rotate to navigate between objects (cost $7.0$).
    On the right, by planning several steps ahead, it finds an efficient path by rotating and pushing stack of books (total cost $23.5$). It avoids a simpler but high-cost detour that only considers Move (cost $50.0$).}
    \label{fig:planner_decision}
    \vspace{-12pt}
\end{figure}

A* search evaluates each state $s$ using the cost function $f(s) = g(s) + h(s)$, where $g(s)$ is the accumulated path cost from the start to state $s$, and $h(s)$ is a heuristic estimate of the remaining cost to the goal. We use the standard Euclidean distance to the target object as the heuristic $h(s)$.
The accumulated path cost $g(s')$ is updated at each step with action $a:s=(p,r,\mathcal{D}) \rightarrow s'=(p',r',\mathcal{D'})$ as:
\begin{equation}
\label{eq:update_cost}
g(s') = g(s) + C(a) + P(s'),
\end{equation}
where $C(a)$ is the action cost and  $P(s')$ is a gripper placement penalty. We define action costs based on the primitive.
Move assigns a cost proportional to the Euclidean distance between two end-effector positions. Rotate applies a fixed cost for orientation changes. Push generates its cost from the value $M'[x,y]$ at the contact point $(x,y)$. 
The penalty $P(s')$ consists of three terms: (i) the average cost of the region the gripper overlaps, (ii) a collision penalty if the gripper collides with an object, and (iii) a proximity penalty proportional to the distance to high-cost objects. 
For additional details of push analysis and the planner, please refer to the appendix. 
By combining these costs, the planner can find a trajectory that minimizes risk while permitting acceptable contact when necessary; see Fig.~\ref{fig:planner_decision} for more information about the A* planner. The path is then converted back to 3D space for evaluation and execution.

\section{Simulation Experiments}
\label{sec:sim_exps}

\subsection{Experiment Setup in Simulation}
\label{ssec:sim_setup}

We build and test our pipeline using PyBullet simulation~\cite{coumans2019}. 
In simulation, we create 20 scenes designed to test contact-rich manipulation. These use a hybrid object dataset, which includes tall and bulky items (\eg sugar boxes and water pitchers) from the YCB~\cite{calli2015ycb} dataset and fragile objects (\eg wine glasses and stacked bowls) using 3D models generated from TRELLIS~\cite{xiang2024structured}. TRELLIS allows constructing delicate object meshes from single-view images and introduces fragility constraints not present in YCB objects. 
All objects are placed on a tabletop. 

We use three cameras to capture the scene for cost map generation: one in front of the scene and two on the sides. The camera in front also captures the scene image $I$ used as input to the VLM. To prevent interactions with the tabletop, we explicitly assign its cost to be 10. 
To quantify the benefit of using VLM-generated costs, we compare with two control groups: (i) with no push action allowed to obtain a collision-free path, which we test in our ``Collision-Free Planning'' baseline (see Sec.~\ref{ssec:baselines_in_sim}),
and (ii) with the cost of all objects set to 0 to see how the planner will plan the path if all collisions are allowed (see Sec.~\ref{ssec:ablation}).

\begin{table*}[t]
    \centering
    \footnotesize
    \begin{tabular}{llccccc}
        \toprule
        \textbf{Category} & \textbf{Path Planning} & \textbf{Reach} & \textbf{Path} & \textbf{Contact} & \textbf{Unsafe Object}  & \textbf{Success}\\
        & \textbf{Algorithm} & \textbf{Target $\uparrow$} & \textbf{Cost $\downarrow$} & \textbf{Duration (s) $\downarrow$} & \textbf{Displacement (cm) $\downarrow$} & \textbf{Rate $\uparrow$} \\
        \midrule
        
        \multirow{4}{*}{Collision\ Free} 
            & RRT      & 66.75\% & - & \textbf{1.12} & 6.47  & 28.75\% \\
            & RRT*     & 61.00\% & - & 1.25 & 7.78  & 26.50\% \\
            & A* w/o direction & 20.00\% & - & 8.33 & 6.51  & 15.00\% \\
            & A*       & 23.33\% & - & 5.14 & \textbf{1.93}  & 15.75\% \\
        \midrule
        \multirow{4}{*}{\baseline}
            & RRT       & 62.91\% & 9.04  & 1.22 & 7.84 & 57.25\% \\
            & RRT*      & 59.50\% & 9.11  & 1.26 & 7.80 & 50.75\% \\
            & A* w/o direction & 58.25\% & 6.71  & 5.93 & 3.82 & 57.50\% \\
            & \textbf{A* (\method)}     & \textbf{78.00\%} & \textbf{5.93}  & 4.18 & 2.51 & \textbf{73.75\%} \\
        \midrule
           - & LAPP  &  50.00\%  & - & 9.37 & 9.22  &  25.00\%\\
        \bottomrule
    \end{tabular}
    \caption{
       Comparison of path planning algorithms and results in PyBullet simulation. We report five quantitative metrics (see Sec.~\ref{ssec:eval_metrics}).
       ``Success Rate" only counts the trajectories that reach the target and satisfy thresholds on ``Path Cost," ``Contact Duration," and ``Unsafe Object (human-designated) Displacement." 
       The arrow $\uparrow$ indicates larger values of the metric correspond to better performance, and $\downarrow$ represents the opposite. Collision-free baselines and LAPP~\cite{xielanguage} do not use VLMs to generate object costs, so they do not have values for ``Path Cost" in this table. 
    }
    \vspace{-20pt}
    \label{tab:simulation}
\end{table*}

\subsection{Motion Planning Methods}
\label{ssec:planning_method}

As presented in Sec.~\ref{ssec:motion_planning_contacts}, \method performs directional contact-aware path planning on the flattened cost map $M'$. 
In simulation, we also test the 3D cost map $V$ with RRT~\cite{lavalle1998rapidly} and RRT*~\cite{karaman2011sampling} as the motion planners, and thus call the respective methods \baseline{}+RRT and \baseline{}+RRT*. Each trajectory consists of a sequence of robot end-effector positions. The total cost of a trajectory $\tau$ is the sum of costs of obstacles the robot arm collides with, ${Cost(\tau) = \sum_{i=1}^{|\tau|} V[\tau_i]}$. Here, $|\tau|$ denotes the length of the trajectory $\tau$, and $\tau_i$ is the position of the $i$-th waypoint. 
In addition, we test another planning method that directly applies A* search without any directional push safety analysis. We refer to it as \baseline{}+A* w/o direction. It only relies on the isotropic costs in the flattened cost map $M$.

\subsection{Baseline Methods}
\label{ssec:baselines_in_sim}

We evaluate \method against the following baselines.

\subsubsection{Collision-Free Planning}
This avoids all collisions, and serves as a baseline to demonstrate that in cluttered environments, a collision-free path may not exist. We test this with A*, A* w/o direction, RRT, and RRT*.

\subsubsection{Language-Conditioned Path Planning (LAPP)} 
We use LAPP~\cite{xielanguage} as a strong baseline, because (like \method) it also allows robots to make collisions with specific objects in the environment. LAPP trains a language-conditioned collision function that predicts whether a robot will collide with objects \emph{other} than the one specified in a language instruction (which the robot is permitted to collide with). The collision function has three inputs: (i) CLIP~\cite{radford2021learningtransferablevisualmodels} image embeddings of the scene, (ii) CLIP text embeddings of the language instruction (e.g., ``can collide with toys''), and (iii) joint configurations of the robot arm. 
We build directly upon the official open-source LAPP code.

\subsection{Evaluation Metrics}
\label{ssec:eval_metrics}

To evaluate the quality of trajectories in our densely cluttered environments, we compute the following metrics in 3D space. 
The robot's trajectory terminates when the robot's end-effector reaches the target, or if a time limit is reached.
\begin{itemize}
    \item \texttt{reach\_target}: whether the robot reaches the target object at the end of the trajectory:
    \[
    \texttt{reach\_target} = \|\mathbf{p}'_e - \mathbf{p}'_{o_{\rm targ}}\| < 0.01 \text{m}.
    \]
    \item \texttt{path\_cost}: the sum of VLM-assigned costs of all collided obstacles during execution:
    \[
    \texttt{path\_cost} = \sum\nolimits_{o \in \mathcal{O} \setminus o_{\text{targ}}} c_o.
    \]
    \item \texttt{contact\_duration}: the sum of duration for which the robot is in contact with any obstacle:
    \[
    \texttt{contact\_duration} = \sum\nolimits_{o \in \mathcal{O} \setminus o_{\text{targ}}} t_o.
    \]
    \item \texttt{displacement}: the displacement of each object:
    \[
    \texttt{displacement}_{o} = \|\mathbf{p}'_o - \mathbf{p}_o\|, o\in \mathcal{O}.
    \]
\end{itemize}
Here, $e$ is the robot end effector and $\mathbf{p}'_e$ is its final 3D position. For each object $o \in \mathcal{O}$, $\mathbf{p}_o$ and $\mathbf{p}'_o$ denote its initial and final 3D position, respectively. The cost of object $o$ is represented by $c_o$ while $t_o$ is the contact duration between the robot and $o$.
After calculating these metrics, we define a trajectory as a success if the following are all true: (i) the robot reaches the target with \texttt{path\_cost} $< 10$, (ii) \texttt{contact\_duration} $< 100\text{s}$ and (iii) for all unsafe obstacles, we have \texttt{displacement} $< 10\text{cm}$. To decide on which objects are unsafe in a given scene, a skilled human annotator pre-selects the 1-2 highest-cost objects in a scene, and those are set for all methods evaluated.

\begin{figure}[t]
    \centering
    \includegraphics[width=0.48\textwidth]{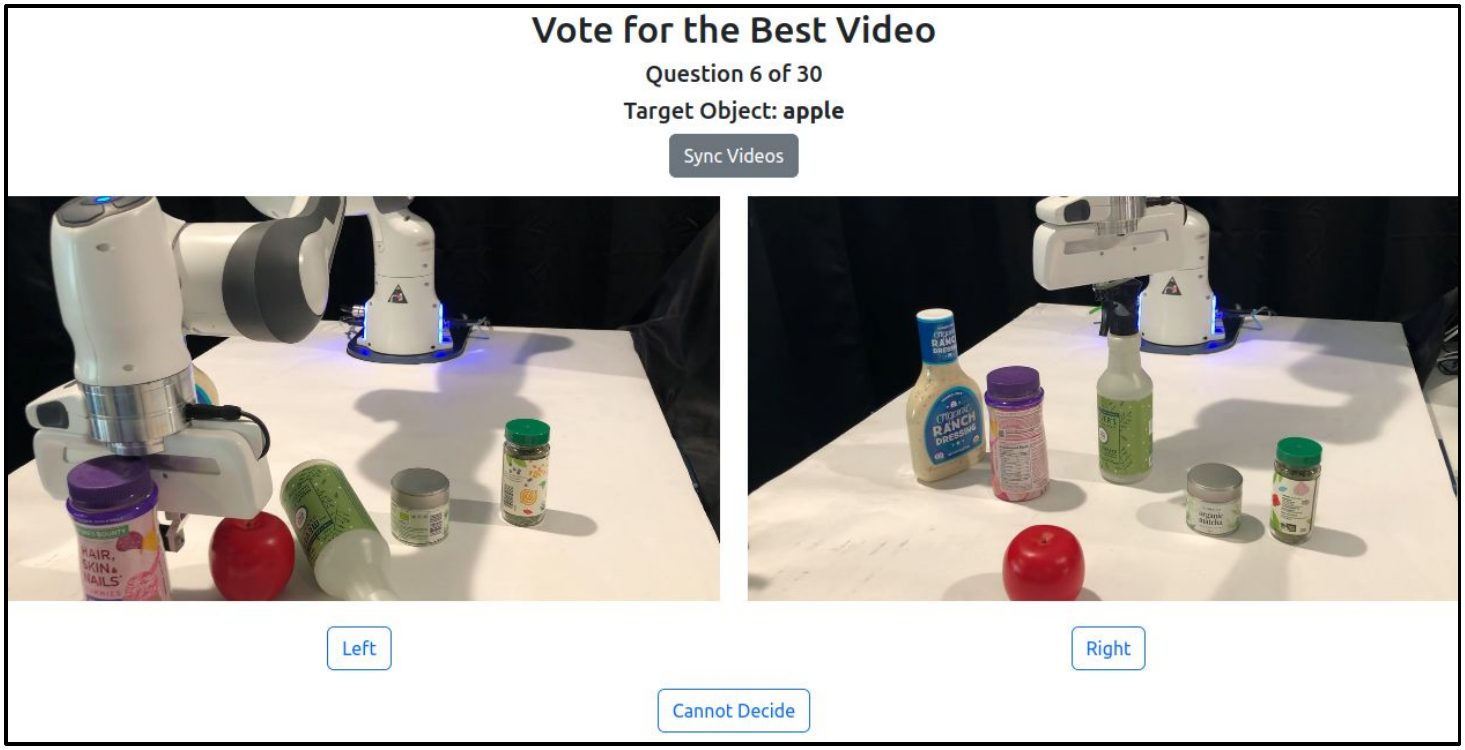}
    \caption{
        Our user study evaluation website interface. For each question, the human evaluates two videos of robot trajectories without knowing the underlying robotics method that caused each robot motion. For each video pair, they select which video is more preferable to them. To aid comparisons, we enable the users to sync the videos. We also allow the option of ``Cannot Decide.''
    }
    \vspace{-15pt}
    \label{fig:human_eval_web}
\end{figure}

\begin{figure*}[t]
    \centering
    \includegraphics[width=\textwidth]{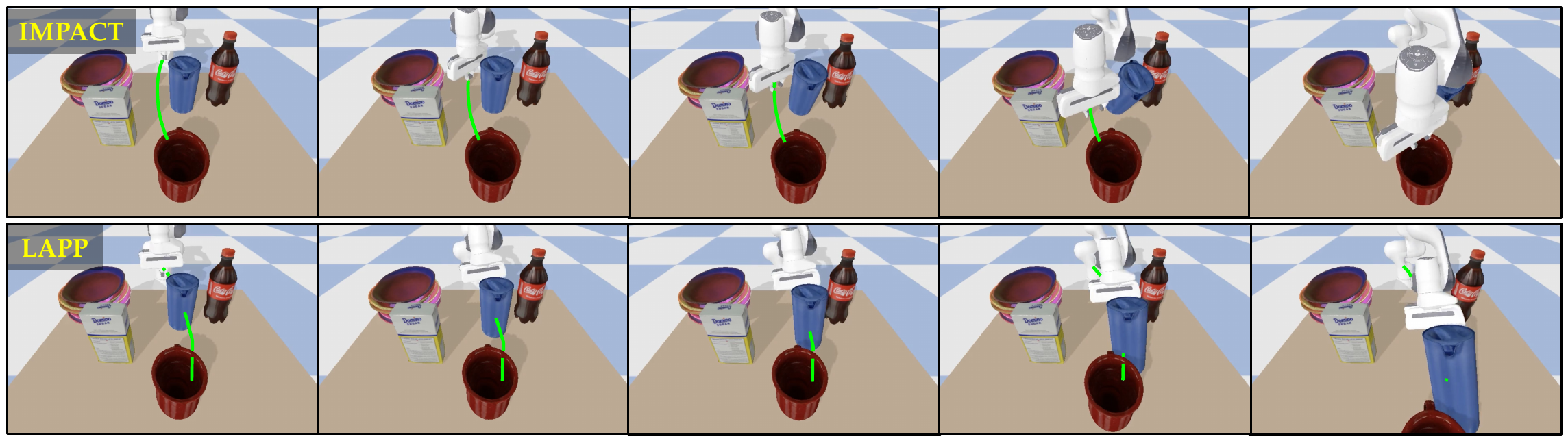}
    \caption{
        Examples of trajectories planned by \method (top row) and LAPP (bottom row) in PyBullet simulation~\cite{coumans2019}. The obstacles are: a coke bottle, a pitcher, a sugar box and a pile of bowls. The target object is the mug behind the obstacles. The planned paths are shown in an overlaid green curve in each image. We also provide LAPP with a language instruction ``Can collide with the pitcher and the sugar box." See Sec.~\ref{ssec:simulation_results} for more details. 
    }
    \label{fig:sim}
    \vspace{-10pt}
\end{figure*}

\subsubsection{User Study Evaluation} 
While the prior quantitative metrics evaluate trajectory cost, they may not fully capture whether a trajectory is ``acceptable'' to humans. 
For example, if all collisions with objects are counted and penalized equally, a trajectory where the robot solidly collides with only one object can have a lower cost compared to one where it gently contacts multiple objects. 
Thus, motion planning algorithms optimize to select the former trajectory, even though the latter trajectory may be more acceptable to humans due to the gentle contacts. 
Furthermore, tolerance for contact may vary from person to person. 

To better assess \method's ability to generate semantically acceptable behavior, we conduct a user study approved by the IRB office of our institution. We develop a website to collect human feedback, where participants evaluate robot trajectories by watching videos (see Fig.~\ref{fig:human_eval_web}). Each question presents two videos of robot trajectories in the same scene with the same target object. One video is generated by an algorithm that utilizes VLM-assigned costs, and the other by the corresponding Collision-Free baseline with the same planner or by LAPP. Users are not informed which algorithm produced each trajectory. Following a similar approach as Mirjalili~et~al.~\cite{mirjalili2023lan}, participants are asked to select the trajectory they find most acceptable. 

\subsection{Simulation Results}
\label{ssec:simulation_results}

Table~\ref{tab:simulation} reports our simulation results. \method achieves the highest success rate with lower path cost, shorter contact duration, and smaller displacement of unsafe objects (pre-selected by human). 
Paths planned by Collision-Free baselines can also lead to unexpected collisions. These baselines prioritize avoiding all obstacles, but the robot may lack continuous feasible joint configurations to execute the path and becomes too close to obstacles, increasing the risk of collisions. 
\method also outperforms the \baseline{}+A* w/o direction, which we attribute to the contact-aware information embedded in the anisotropic cost map.
The LAPP baseline has a lower \texttt{reach\_target} and success rate, mainly because it sometimes fails to find paths or push obstacles along the path.

\begin{figure}[t]
    \centering
    \includegraphics[width=0.49\textwidth]{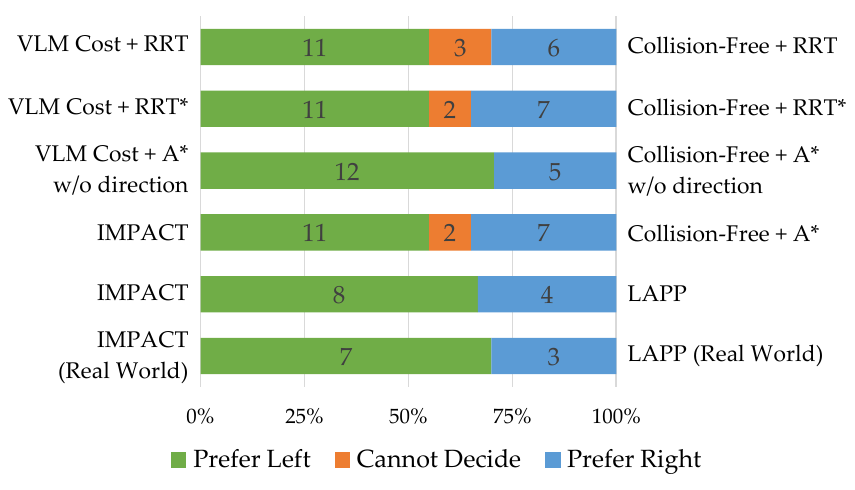}
    \caption{
        Human evaluation results. The first five rows report results from simulation, and the last row show the real-world experiments. Each bar represents the results of comparing two methods. It is divided into three segments: the left method, the right method, and ``Cannot decide." Each segment counts the number of questions where people prefer the trajectory generated by the corresponding method. For example, in the top row (\baseline{}+RRT vs. Collision-Free+RRT), more participants prefer the trajectory planned by \baseline{}+RRT in 11 cases out of 20, Collision-Free+RRT in six cases, and have no preference in three cases.
    }
    \label{fig:human_eval}
    \vspace{-10pt}
\end{figure}

Fig.~\ref{fig:sim} shows trajectories planned by different methods in the same scene. \method finds a trajectory that makes contact with the pitcher. 
LAPP also guides the robot to make contact with the pitcher while avoiding the pile of bowls and the coke bottle, but the pitcher is pushed along the path, preventing the robot from reaching the target. As shown in Table~\ref{tab:simulation}, this common behavior leads to a higher average unsafe object \texttt{displacement}. 

Fig.~\ref{fig:human_eval} shows the human evaluation results collected from 25 participants in total. We collect 100 video pairs in simulation and uniformly sample 20 questions for each participant. Each question contrasts a trajectory generated by one method against some baseline. 
The results show that the method using VLM-assigned costs is the most preferred method across all scenes, suggesting that cost maps produce motion plans that better align with human preferences by leveraging commonsense knowledge in the VLM.

\subsection{Ablation Study}
\label{ssec:ablation}

To investigate the benefit of VLM-generated object costs, we conduct an ablation study by setting all object costs to 0, so all collisions are allowed during path planning. Results in Table~\ref{tab:simulation_ablation} demonstrate that VLM-assigned costs improve the trajectory success rate (see Sec.~\ref{ssec:eval_metrics}) of reaching tasks. For all three motion planning algorithms, the VLM takes properties of obstacles into account and assigns object costs based on different scenes. This encourages planners to avoid obstacles and generate trajectories with reduced contact. 

\begin{table}[t]
    \centering
    \footnotesize
    \begin{tabular}{llc}
        \toprule
        \textbf{Path Planning} & \textbf{Cost} & \textbf{Success Rate}\\
        \textbf{Algorithm} & &   \\
        \midrule
        \multirow{2}{*}{RRT} 
            & \baseline & \textbf{57.25\%} \\
            & Same Cost for All & 44.50\% \\
        \midrule
        \multirow{2}{*}{RRT*} 
            & \baseline    & \textbf{50.75\%} \\
            & Same Cost for All & 42.50\% \\
        \midrule
        \multirow{2}{*}{A*} 
            & \baseline (\method)   & \textbf{73.75\%} \\
            & Same Cost for All & 65.00\% \\
        \bottomrule
    \end{tabular}
    \caption{
       Results of our ablation study on object costs. ``Same cost for all" refers to assigning all object costs to be 0, instead of querying the VLM to generate costs. See Sec.~\ref{ssec:ablation} for details. 
    }
    \vspace{-8pt}
    \label{tab:simulation_ablation}
\end{table}


\begin{figure*}[t]
    \centering
    \includegraphics[width=\textwidth]{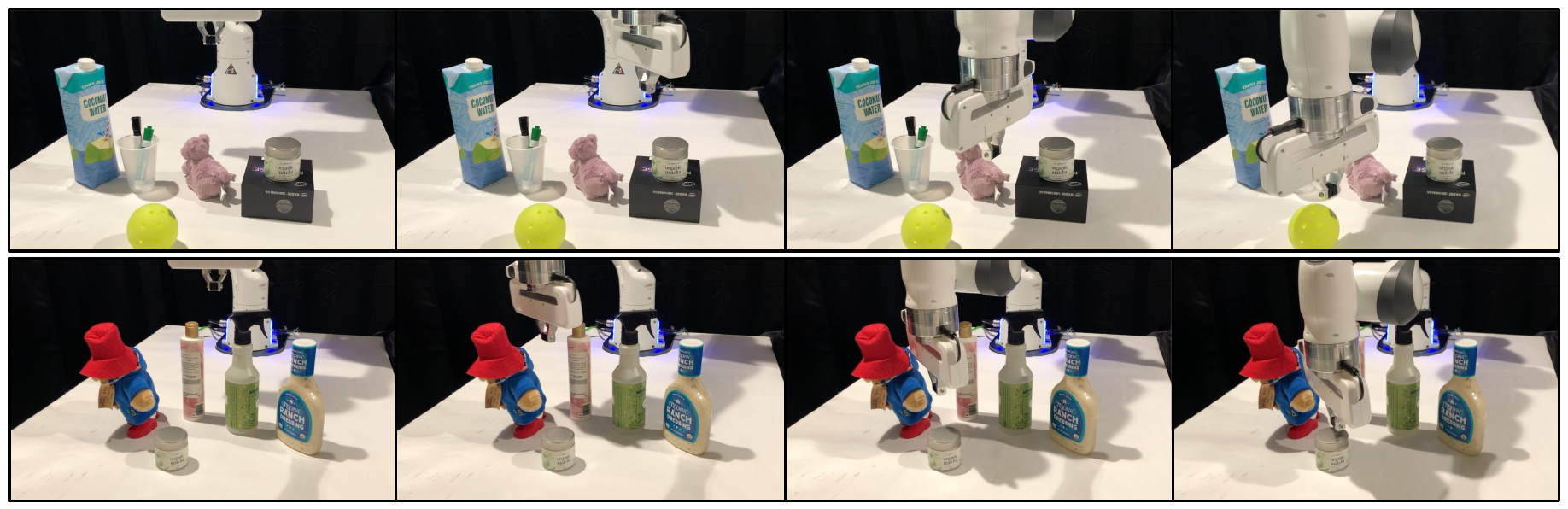}
    \caption{Examples of successful trajectories from \method in the real world. Top row: the robot reaches the target pickleball while moving closer to the soft toy hippo. Bottom row, the robot rotates its gripper to avoid contact with tall bottles and reaches the target matcha can.
    }
    \vspace{-15pt}
    \label{fig:experiments_real}
\end{figure*}

\section{Physical Experiments}
\label{sec:physical_exps}

\subsection{Experiment Setup}
\label{ssec:physical_setup}

\begin{table}[t]
    \centering
    \footnotesize
    \begin{tabular}{lcc}
        \toprule
        \textbf{Method} & \multicolumn{2}{c}{\textbf{Success Rate}}\\
        \midrule
        \method & \multicolumn{2}{c}{\textbf{61\%}}\\
        LAPP & 49\% (Seen Obj.) & 40\% (Unseen Obj.)\\
        \bottomrule
    \end{tabular}
    \caption{
        Results of our method versus LAPP in the real world. Seen objects refer to objects seen during fine-tuning of LAPP while unseen objects are novel objects to LAPP. Our method is zero-shot, so all the objects can be considered unseen to it.
    }
    \vspace{-15pt}
    \label{tab:real}
\end{table}

We evaluate \method on a real robotic system to validate our findings from simulation. This does not involve sim-to-real transfer, as we provide real-world images directly to the VLM for \method. The hardware setup consists of a Franka Panda arm with a standard parallel-jaw gripper, and a flat tabletop surface.
A human operator arranges multiple objects in close proximity on top of the surface. 
For reasonably fair comparisons among methods, the human tries to place objects in consistent locations for each scene. 

We mount two Intel RealSense D435 cameras on both sides of the scene to capture RGBD images and generate voxel grids. 
In the real world, we need to get each object's position to assign the associated object cost to those voxels. Unlike in simulation, we use Grounded SAM 2~\cite{ravi2024sam,ren2024grounded} to generate the segmented point cloud of each object as we build the cost map. 
Each image and the object list are provided as input to Grounded SAM 2 to predict the segmentation masks of the objects. We then get a multi-view segmented point cloud and convert it to the cost map. The motion planning part is the same as in simulation.

For physical experiments, we test \method and LAPP. 
To adapt LAPP for real-world experiments, we fine-tune it on a small dataset containing 25 real-world RGB images, manually annotated with collision scores. The images are captured by an Intel RealSense D435 camera that faces both the robot and the scene. The annotations explicitly label which objects are safe to collide with (for example, ``can collide with plastic bottle''). Objects not mentioned in the language prompts are treated as unsafe by default, aligning with LAPP’s methodology of explicit collision rules. Human operators pair each scene with joint configurations and task-aligned prompts, reflecting the original work’s use of free-form language constraints~\cite{xielanguage}. This retains LAPP’s pre-trained reasoning while grounding predictions in real-world spatial relationships and safety priorities. Of the 10 test scenes, 8 use objects seen during fine-tuning, while 2 scene contains novel objects to evaluate generalization.

\subsection{Real World Results}

Table~\ref{tab:real} shows real world results of different methods. \method outperforms LAPP on average across all scenes. Furthermore, the performance gap is larger in scenes with unseen objects (61\% versus 40\% success).
The results suggest the strong generalization ability of our method since it does not need sim-to-real transfer. However, LAPP requires fine-tuning on unseen real world objects to achieve comparable results.
We also conduct a user study on real-world experiments. The same 25 participants evaluate 10 pairs of real-world trajectories. As shown in the last row of Fig.~\ref{fig:human_eval}, \method is preferred in more scenarios. Fig.~\ref{fig:experiments_real} shows two examples of successful trajectories planned by \method.

\begin{figure}[t]
    \centering
    \includegraphics[width=0.47\textwidth]{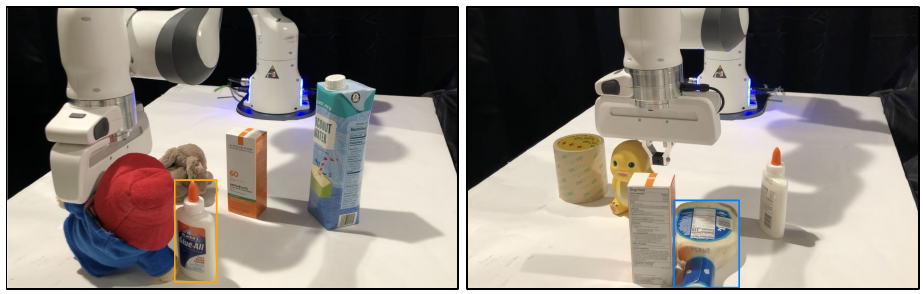}
    \caption{Examples of real-world failures generated by \method. In the left image, the robot gets stuck on the toy bear while reaching the \textcolor{orange}{target glue bottle}. In the right image, the robot does not rotate the gripper in time and collides with the \textcolor{lightblue}{ranch bottle}.}
    \label{fig:failure}
    \vspace{-15pt}
\end{figure}

\textbf{Failure Cases}. While \method generates semantically-acceptable trajectories in most trials, failures can happen during execution. Fig.~\ref{fig:failure} shows two examples of failed trajectories. The main failure cases are: (i) part of the robot gets stuck on an obstacle, (ii) the gripper collides with an obstacle before or during rotation, and (iii) the VLM predicts object costs that are misaligned with human preferences.

\section{Limitations}
\label{sec:limitations}

While promising, \method has several limitations that point to interesting directions for future work. 
First, after selecting the most semantically-acceptable trajectory, the robot follows it open-loop. This means it cannot react to unexpected disturbances in real-time. 
Second, \method relies on having relatively complete RGBD observations, and may be less effective under partial observability with severe occlusions. 
Developing a closed-loop procedure that can actively perceive the environment may address these limitations.


\section{Conclusion}
\label{sec:conclusion}

In this work, we introduce \method, a framework for motion planning in clutter that leverages the broad knowledge in Vision-Language Models (VLMs) to assess object contact tolerance. \method represents this information in an anisotropic cost map for motion planning. Our results in simulation and in real-world experiments demonstrate that robots can efficiently reach targets while making semantically-acceptable contact when needed. We hope that this work inspires future work towards flexible and contact-rich robot manipulation in densely cluttered environments.

\section*{ACKNOWLEDGMENTS}

{\small
We thank our colleagues Ebonee Davis, Sicheng He, Ayano Hiranaka, Minjune Hwang, Yunshuang Li, and Qian (Peter) Wang for helpful technical advice and discussions. 
}

\renewcommand{\biblabelsep}{0.5em}  
\printbibliography

\clearpage
\section*{APPENDIX}

\subsection{Example Simulation Setup and Cost Map Generation}

\begin{figure}[t]
    \centering
    \includegraphics[width=0.48\textwidth]{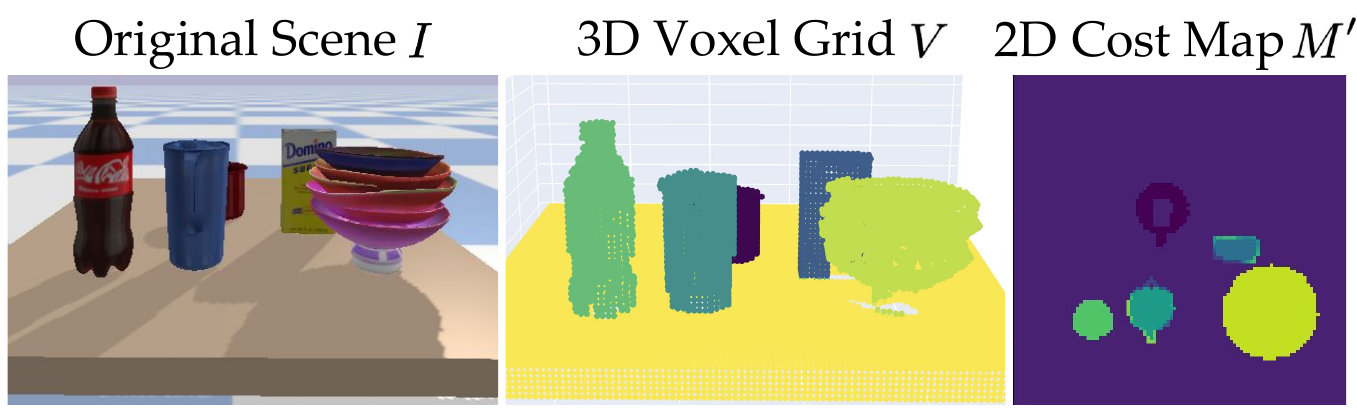}
    \caption{
        Our simulation setup and in PyBullet simulation~\cite{coumans2019}. In this example scene, there is a coke bottle, a pitcher, a mug, a sugar box and a pile of bowls on the tabletop. The objective is to reach the mug. The 3D voxel grid $V$ is generated from the VLM-assigned costs, updated during push outcome analysis, and ultimately transformed into the final anisotropic cost map $M'$. The yellow and green regions (e.g., for the coke bottle and the pile of bowls) indicate higher costs. 
        For visual clarity, we show the original scene $I$, before it is annotated for the VLM. 
    }
    \label{fig:sim_setup}
\end{figure}

Fig.~\ref{fig:sim_setup} shows an example of the physical environment in simulation, and its corresponding 3D voxel grid $V$ and 2D anisotropic cost map $M'$.

\subsection{Additional Details of Push Outcome Analysis}

As defined in Eq.~\ref{eq:safety_score} in Sec.~\ref{sssec:cost_map_gen}, the aggregated safety score $f_s(x,y)$ of push outcomes at position $(x,y)$ depends on two components: a probabilistic likelihood function $l$ and categorical safety score $u$. 

The probabilistic model $l$ quantifies the deviation of a sampled outcome from the object's reverse surface normal. For the $i$-th push outcome of $m$ samples, its likelihood is computed using a 2D Gaussian decay function:
\begin{equation}
l_i(\delta_i, \theta_i)= \exp \left(-\left(\left(\frac{\delta_i}{\sigma_\delta}\right)^2+\left(\frac{\theta_i}{\sigma_\theta}\right)^2\right)\right),
\end{equation}
where $\delta_i$, $\theta_i$ are the variations in distance and angle, and $\sigma_\delta$, $\sigma_\theta$ are standard deviation parameters. We uniformly random sample $\delta_i\in [20, 50]$ and $\theta_i\in [-25, 25]$. The parameters are set to $\sigma_\delta=33, \sigma_\theta=25$.

The safety score $u$ assigned to each push outcome is discrete and based on the outcome category. Here, the push outcome refers to the collision status between the pushed object and all other objects, including the target. These scores are chosen to reflect the risk of damage according to different types of collision in the resulting scene. 
Table~\ref{tab:push_outcome} shows the details of the four categories. Low scores assigned to high-risk pushes discourage the planner to select contact points and push directions that would lead to unsafe scenes or unintended collisions.

\subsection{Additional Details of Contact-Aware Planner}

The action costs of three motion primitives in Sec.~\ref{sssec:a_star_planning} are defined as follows:
\begin{equation}
C(a) =
\begin{cases}
    \lambda_m \cdot d(p,p') & \text{if } a = \text{Move}, \\
    \lambda_r & \text{if } a = \text{Rotate}, \\
    \lambda_p +\lambda_s M'[x,y] & \text{if } a = \text{Push}.
\end{cases}
\end{equation}
Here, $\lambda_m,\lambda_r,\lambda_p, \lambda_s$ are positive constants, $d(\cdot,\cdot)$ is the Euclidean distance between the initial and final end-effector positions, and $(x,y)$ is the contact position between the end-effector and the object during the push. In our experiments, we set $\lambda_m=1,\lambda_r=1.8,\lambda_p=8, \lambda_s=10$.

This action cost function is designed to encode the objectives of minimal-contact motion and risk-aware interaction with the environment. For the Move primitive, the cost scales linearly with the distance of the end-effector, which aligns with the principle that shorter motions usually have lower risk and lower cost. The cost of the Rotate primitive is a constant. Since Rotate only alters the end-effector’s orientation without changing its position and does not disturb the environment in most cases, a constant cost discourages unnecessary reorientations while still allowing them when beneficial. In particular, rotations often enables safer or more effective pushes. The action cost of the Push primitive consists of the baseline effort of initiating contact and the directional cost of the interaction embedded in the anisotropic cost map $M'$. The function ensures that pushes are executed selectively and lead to safer contact.

For each state $s$ during A* planning, the gripper placement penalty $P(s)$ introduced in Eq.~\ref{eq:update_cost} is a critical component of the accumulated path cost. It promotes the gripper placement validity and safety by penalizing undesirable end-effector poses.
A placement is invalid if the gripper leaves the robot workspace, intersects a high-cost object or contains any voxel with a cost exceeding the predefined maximum threshold in $M'$.
If the gripper placement passes these strict checks, we calculate the penalty as the sum of three weighted terms: (i) the average cost of the voxels in $M'$ that the gripper overlaps, which guides the robot to enter lower-cost regions, (ii) an object contact penalty when the gripper makes contact with low-cost objects, and (iii) a proximity penalty, which grows when the gripper is closer to high-cost obstacles. These three penalty terms measure the quality of a gripper placement and ensure that the planner prefers stable and collision-free gripper poses that are away from higher-cost voxels.


\begin{table}[t]
    \centering
    \footnotesize
    \begin{tabular}{lc}
        \toprule
        \textbf{Push Outcome Category} & \textbf{Safety Score}  \\
        \midrule
        Safe & 1.000 \\
        \midrule
        Low-cost contact & 0.600 \\
        \midrule
        High-cost contact & 0.100 \\
        \midrule
        Contact with the target & 0.025 \\
        \bottomrule
    \end{tabular}
    \caption{
        Safety score of different push outcomes. Higher scores are assigned to safe pushes (the pushed object does not collide with others) and collisions with low-cost objects, while collisions with high-cost objects or even the target have low scores. 
    }
    \vspace{-8pt}
    \label{tab:push_outcome}
\end{table}




\end{document}